\documentclass[conference]{IEEEtran}
\IEEEoverridecommandlockouts
\usepackage{cite}
\usepackage{amsmath,amssymb,amsfonts}
\usepackage{algorithmic}
\usepackage{graphicx}
\usepackage{textcomp}
\usepackage{xcolor}
\usepackage{url} 
\usepackage{orcidlink}

\usepackage{subcaption}

\usepackage{multirow}  
\usepackage{array}     
\usepackage{booktabs}

\usepackage{textcomp} 

\def\BibTeX{{\rm B\kern-.05em{\sc i\kern-.025em b}\kern-.08em
    T\kern-.1667em\lower.7ex\hbox{E}\kern-.125emX}}
\begin{document}


\IEEEoverridecommandlockouts
\IEEEpubid{\begin{minipage}[t]{\textwidth}\ \\[10pt]
        \raggedright\normalsize\color{gray}{979-8-3315-0969-9/25/\$31.00 \copyright2025 European Union \hfill}
\end{minipage}} 

\title{Automated Pollen Recognition in Optical and Holographic Microscopy Images\\}


\author{
\IEEEauthorblockN{
Swarn Singh Warshaneyan\orcidlink{0000-0003-3815-6329}\IEEEauthorrefmark{2}\textsuperscript{*}, 
Maksims Ivanovs\orcidlink{0000-0003-2477-7327}\IEEEauthorrefmark{2}, 
Blaž Cugmas\orcidlink{0000-0002-3615-7443}\IEEEauthorrefmark{3}\IEEEauthorrefmark{4}, 
Inese Bērziņa\IEEEauthorrefmark{6}\IEEEauthorrefmark{4}, \\
Laura Goldberga\IEEEauthorrefmark{4}, 
Mindaugas Tamosiunas\orcidlink{0000-0001-5866-9557}\IEEEauthorrefmark{4}, 
Roberts Kadiķis\orcidlink{0000-0001-6845-4381}\IEEEauthorrefmark{2}
}
\IEEEauthorblockA{
\IEEEauthorrefmark{2}Institute of Electronics and Computer Science, Riga, Latvia \\
\IEEEauthorrefmark{3}Vetamplify SIA, Riga, Latvia \\
\IEEEauthorrefmark{4}University of Latvia, Riga, Latvia \\
\IEEEauthorrefmark{6}VetCyto SIA, Riga, Latvia
}
\IEEEauthorblockA{
\textsuperscript{*}Correspondence: swarn.warshaneyan@edi.lv
}
}

\maketitle

\begin{abstract}

This study explores the application of deep learning to improve and automate pollen grain detection and classification in both optical and holographic microscopy images, with a particular focus on veterinary cytology use cases. We used YOLOv8s for object detection and MobileNetV3L for the classification task, evaluating their performance across imaging modalities. The models achieved 91.3\% mAP50 for detection and 97\% overall accuracy for classification on optical images, whereas the initial performance on greyscale holographic images was substantially lower. We addressed the performance gap issue through dataset expansion using automated labeling and bounding box area enlargement. These techniques, applied to holographic images, improved detection performance from 2.49\% to 13.3\% mAP50 and classification performance from 42\% to 54\%. Our work demonstrates that, at least for image classification tasks, it is possible to pair deep learning techniques with cost-effective lensless digital holographic microscopy devices.

\end{abstract}

\begin{IEEEkeywords}
automated pollen recognition, machine learning, holographic microscopy, medical imaging, veterinary medicine
\end{IEEEkeywords}

\section{INTRODUCTION}

Microscopy is an integral part of most veterinary medicine diagnostic procedures. The use of optical microscopes with complex lens configurations to obtain colored (RGB) images of relevant biological structures has become the golden standard of diagnostic imaging. Despite that, the various costs and complexities that establishing a microscopic diagnostic facility involves have made such services inaccessible in poorly connected and less developed areas. Recent technological advancements have led to the development of lensless digital holographic microscopy (DHM) \cite{b1}, which is capable of becoming a highly cost-effective and much less complicated alternative to conventional optical microscopy, making veterinary services more accessible and improving care for animals in need. 

Similar to many other common microscopic particles, pollen grains are constantly present in the atmosphere and are involved in several important environmental and biological phenomena such as seasonal allergies, spread of species, changes in weather, crop yields, natural vegetation patterns, and climate change \cite{b2}. Therefore, developing an affordable and accessible system for recognizing diverse pollen types can support professionals working in related fields. In particular, such a system can assist in veterinary diagnostics of many different health conditions that are associated with exposure to certain types of pollen cells, such as adverse immune responses caused by overexposure. 

Taking into account the importance of pollen grain recognition, it appears promising to use machine learning (ML) methods for automating it. Furthermore, such applications of ML to the processing of lensless DHM images has the potential to benefit not only veterinary medicine but also microscopy imaging in general. In particular, it can enable greater adoption of microscopy techniques by significantly reducing overall costs and increasing the convenience of using this technology, thus benefiting communities affected by environmental pollen exposure \cite{b3}. Therefore, the goal of this study was to evaluate and improve the performance of selected deep learning models for object detection and classification on optical and holographic microscopy images, with a view towards possible deployment on edge devices. A targeted study involving scans of sample slides containing pollen grains relevant for veterinary cytology was used as a proxy for similar diagnostic use cases.

\section{RELATED WORK}

The analysis of pollen grain images has benefited from advancements in computer vision and machine learning. Classical computer vision approaches involve algorithmic methods such as feature extraction based on brightness and shape descriptors to characterize pollen grains. Thus, a framework using such descriptors derived from intensity images and related to the ornamentation and morphology of pollen grains was proposed in \cite{b4}. In \cite{b5}, an annotated image dataset for Brazilian Savannah pollen types was presented, and a baseline of human and computer performance was established using various feature extractors and machine learning techniques. More recently, there has been a shift towards growing adoption of deep learning models for the purpose of pollen image classification. Thus, in \cite{b6}, the large annotated image dataset for pollen grains classification POLLEN73S was introduced, and  Convolutional Neural Networks (CNNs) were efficiently applied to this task. In \cite{b7}, a deep learning-based system for automated multifocus pollen detection and classification on microscope slides was developed. The use of pretrained CNNs and transfer learning for efficient pollen grain classification was explored in \cite{b8}. In \cite{b9}, a deep learning model for tree species identification using pollen grain images was developed. Furthermore, in \cite{b10}, deep learning was utilized for single-frame 3D lensless microscopic imaging of pollen grains. Improved pollen grain image classification using deep learning, specifically CNN-based methods, was achieved in \cite{b11}. The automatic classification of pollen grains relevant to the field of veterinary medicine was explored in \cite{b2}, with their model achieving an average accuracy of 88\%, and in a related study \cite{b12}, it was discovered that utilizing silicone instead of adhesive tape as methods for sample fixation on whole slides provides images that are better suited for automated analysis.

Furthermore, recent academic work has increasingly applied  computer vision and deep learning techniques in conjunction with holographic imaging setups. Thus, \cite{b13} used high-throughput submersible imaging with a digital in-line holographic microscope and transfer learning through CNNs for plankton species classification in artificial seawater monocultures and natural seawater samples. In \cite{b14}, deep neural networks, particularly CNNs and Vision Transfomers (ViTs), were used for applying digital holography in the context of 3D computer micro-vision for micro-robotics applications. In \cite{b15}, a method for rapid optical screening of Bacillus anthracis spores using holographic microscopy and deep learning to allow superior diagnosis of similar pathogens was presented. This method relies on a custom-made and specialized CNN for the purpose of classifying holographic images of unlabeled living cells, with optical images used as points of reference. Although \cite{b15} focused on anthrax rather than the veterinary cytological samples targeted in our project, their work is  particularly relevant to our research, as it combined holographic imaging with deep learning for biological sample classification using microscopy images, while similarly using model performance on optical images for comparative analysis.

\section{DATA}

The study was conducted using microscopy images of sample slides containing pollen grains fixed using standard veterinary cytology techniques using adhesive tape. Two separate types of imaging modalities were utilized: RGB optical and greyscale holographic. The optical microscopy images were acquired using a conventional brightfield optical microscope paired with an off-the-shelf digital slide scanner (the Ocus 20 by Fisher Scientific). The holographic microscopy images were acquired using a custom-made lensless DHM \cite{b16}. Overall, the acquired dataset consisted of 20 high-resolution RGB optical microscopy images and 11 high-resolution greyscale holographic microscopy images. The dataset was then used for the deep learning model training, validation and testing as described in the following sections. All of the experimental steps were carried out using the Python programming language in conjunction with the Jupyter development environment. The data splitting was done with the commonly used ratios of 70:15:15 for the training, validation, and testing subsets.

\subsection{Annotation Labels}

The annotation of optical images involved manually creating bounding boxes to identify and classify pollen grains. However, due to the very large number of pollen grains in each image, it was not possible to annotate all of them by hand. Hence, the YOLOv8s model \cite{b17} was trained on a high-quality subset of the manually labelled images for specifically recognizing pollen grains without any regard for their classes to create automatically generated bounding box labels, thus substantially increasing the amount of the annotated data.

The image preprocessing part involved splitting large images into smaller ones; as a consequence, the total number of object instances vary between different types of tasks (i.e, for object detection vs classification), as some instances can be damaged by the splitting. However, even when considering that, the use of automatic labelling substantially increased the size of the dataset each time. In particular, there were 5590 instances of manually labelled pollen grains across 20 optical images for the object detection training step; the use of automated labelling increased that to 77268 instances, which is an increase by a factor of 13.82. There were 4536 instances of manually labelled pollen grains across 20 optical images for the object classification task; the use of automated labelling increased that to 68268 instances, which is an increase by a factor of 15.05. 

Furthermore, there were 2952 instances for manually made labels of pollen grains, obtained from optical images through image alignment (see the following section for details), across 11 holographic images for the object detection task. The use of automated labeling increased that to 71639 instances, which is an increase by a factor of 24.26. There were 2437 instances of manually made labels of pollen grains across 11 holographic images in the object classification training step. The use of automated labeling increased that to 63018 instances, which is an increase by a factor of 25.85.

\begin{figure*}[!htbp]
    \centering
    \begin{subfigure}[b]{0.49\textwidth}
        \centering
        \includegraphics[width=\textwidth]{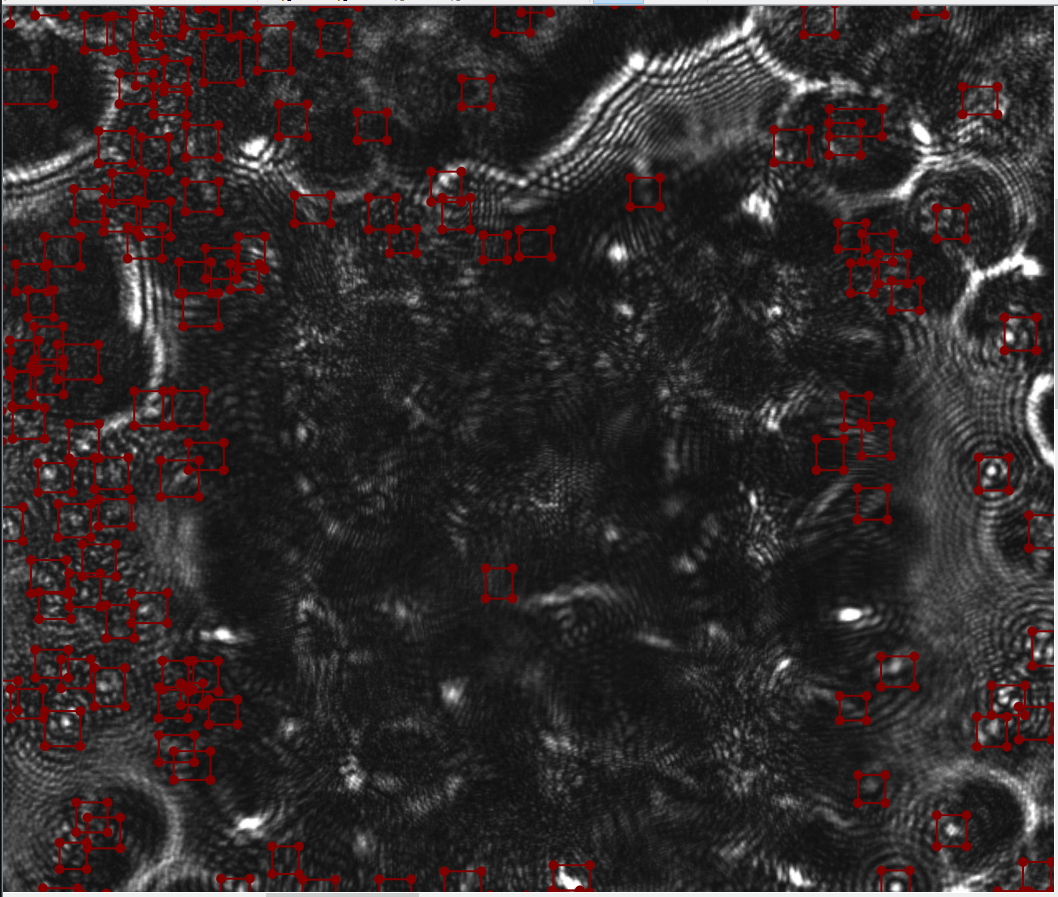}
        \caption{Pollens in an aligned holographic image with automated labels at 40\% zoom and no bounding box expansion.}
        \label{fig:hm-automated-labels}
    \end{subfigure}
    \hfill
    \begin{subfigure}[b]{0.49\textwidth}
        \centering
        \includegraphics[width=\textwidth]{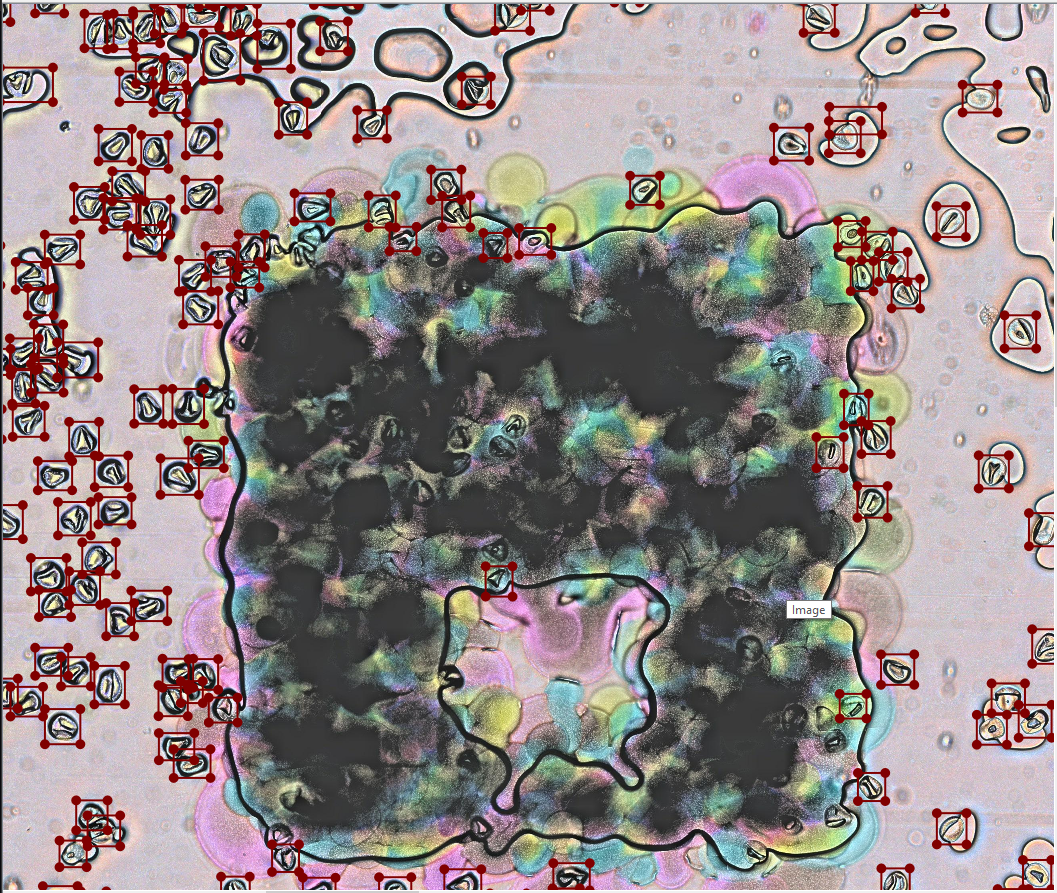}
        \caption{Pollens in an optical image with automated labels at 40\% zoom and no bounding box expansion.}
        \label{fig:om-automated-labels}
    \end{subfigure}
    \caption{Comparing a matched pair of aligned holographic (left) and optical (right) images.}
    \label{fig:comparison-hm-om}
\end{figure*}

\subsection{Image Alignment}

Since it is difficult for humans to identify and classify pollen grains in greyscale holograms, we reused the labelling from the optical modality. The holographic images were aligned with the optical images using image registration methods, making it possible to use the bounding box labels made for the optical images on the holographic images directly without any changes.

The holographic images all had the same dimensions of 3840 by 2160 pixels, with an average file size of 15.8 megabytes. The optical images had slightly differing dimensions averaging to 17500 by 8000 pixels, with an average file size of 55 megabytes. To avoid significant issues due to potential information loss, the holographic images were upsampled through affine transformation to match their larger optical counterparts using annotated points shared between corresponding images and a similarity transformation technique. The image registration was performed as follows:

\begin{enumerate}
\item 	Centroid calculation step involved computing the mean center position of points in both source and destination images.
\item 	Point demeaning step involved subtracting the centroid from each point set.
\item 	Norm calculation involved determining the scale relationship between the two point sets.
\item	Normalization step involved adjusting the differently measured values on a common scale for the demeaned points.
\item	Rotation determination step involved obtaining the values for the optimal rotation matrix through Singular Value Decomposition (SVD).
\item	Transformation matrix construction step involved combining scale, rotation, and translation into a unified transformation.
\end{enumerate}

Due to the difference in dimensions between the images from the two modalities, some purely black areas were introduced at the edges of the holographic images after they were aligned with the optical images. To avoid dataset contamination issues, all aligned holographic images were screened using an automated script after splitting and before training models, and the images with 20\% or more purely black areas were excluded from further experiments. This procedure also contributed to the total number of object instances being slightly different between the two image types in the dataset during the training step.

After such preprocessing, we were able to use the holographic images in the same object detection and classification experiments with YOLOv8s and MobileNetV3L \cite{b18} that we carried out for the optical images.

\subsection{Bias Resolution}

Since the pollen grains belonged to four different species,  the images were divided into four classes. We observed a class imbalance for the holographic images in our dataset, with a substantial gap between the number of object instances belonging to each class. There were 522 manually labeled instances from 4 images for class T1, 105 manually labeled instances from 1 image for class T2, 318 manually labeled instances from 2 images for class T5, and 1516 manually labeled instances from 4 images for class T9. The gap was reduced but not eliminated when using automatic labels, that were available for all of the 20 images. To avoid bias in the models due to the class imbalance during the training and evaluation, we used image augmentation together with other techniques during the training steps.

We applied two mostly similar yet separate approaches involving image augmentation to resolve potential bias issues for the classification and detection training steps, based on the primary difference being the fact that class labels are essential for the classification task but can be ignored for the detection task. After completing all the pre-processing steps necessary to ensure that the image dataset was clean and balanced, we moved on to the next phase in the experimental pipeline.

\begin{table}
\centering
\caption{Number of instances (annotated pollen grains) in each class (separate species) for optical images}
\label{tab:instances}
\begin{tabular*}{\columnwidth}{@{\extracolsep{\fill}}l cc@{}}
\toprule
& \multicolumn{2}{c}{Object Instances} \\
\cmidrule{2-3}
Classes & \textit{Manual Labels} & \textit{Automated Labels} \\
\midrule
T1 & 790 & 7364 \\
T2 & 766 & 7751 \\
T5 & 1288 & 9115 \\
T9 & 1692 & 44038 \\
\midrule
Total & 4536 & 68268 \\
\bottomrule
\end{tabular*}
\end{table}

\begin{table}
\centering
\caption{Data distribution across training, validation, and testing splits for optical images}
\label{tab:data-splits}
\begin{tabular*}{\columnwidth}{@{\extracolsep{\fill}}l cccc@{}}
\toprule
\multirow{2}{*}{Annotation Method} & \multicolumn{4}{c}{Data Splits} \\
\cmidrule{2-5}
 & \textit{Training} & \textit{Validation} & \textit{Testing} & \textit{Total} \\
\midrule
Manual Labels & 3174 & 680 & 682 & 4536 \\
Automated Labels & 47785 & 10241 & 10242 & 68268 \\
\bottomrule
\end{tabular*}
\end{table}

\begin{table}
\centering
\caption{Number of instances (annotated pollen grains) in each class (separate species) for holographic images}
\label{tab:holo-instances}
\begin{tabular*}{\columnwidth}{@{\extracolsep{\fill}}l cc@{}}
\toprule
& \multicolumn{2}{c}{Object Instances} \\
\cmidrule{2-3}
Classes & \textit{Manual Labels} & \textit{Automated Labels} \\
\midrule
T1 & 518 & 6786 \\
T2 & 105 & 7547 \\
T5 & 313 & 8261 \\
T9 & 1501 & 40424 \\
\midrule
Total & 2437 & 63018 \\
\bottomrule
\end{tabular*}
\end{table}

\begin{table}
\centering
\caption{Data distribution across training, validation, and testing splits for holographic images}
\label{tab:data-splits-holo}
\begin{tabular*}{\columnwidth}{@{\extracolsep{\fill}}l cccc@{}}
\toprule
\multirow{2}{*}{Annotation Method} & \multicolumn{4}{c}{Data Splits} \\
\cmidrule{2-5}
 & \textit{Training} & \textit{Validation} & \textit{Testing} & \textit{Total} \\
\midrule
Manual Labels & 1704 & 366 & 367 & 2437 \\
Automated Labels & 44110 & 9453 & 9455 & 63018 \\
\bottomrule
\end{tabular*}
\end{table}

\section{EXPERIMENTS}

The two deep learning models selected for our experiments were YOLOv8s (version 8 Small; \cite{b17}) for object detection and MobileNetV3L (version 3 Large; \cite{b18}) for object classification. Both of these models are suitable for deployment on edge devices due to their state-of-the-art performance on the respective tasks despite their compact size. Although YOLOv8s is capable of performing detection and classification tasks simultaneously, its classification head is architecturally optimized for speed within the detection pipeline, using simplified feature extraction. MobileNetV3L employs dedicated architectural components that are specifically designed to maximize classification accuracy. This architectural specialization makes MobileNet better suited for classification tasks where fine-grained feature discrimination is critical.

We made use of transfer learning and fine-tuning approaches for training the chosen models. After dividing the dataset in three parts using training-validation-testing splitting, we conducted training on the optical images before moving on to the holographic images. Similarly, we conducted training with the manually created labels before moving on to the automatically created labels. Later on, we introduced the expansion of all bounding box areas by 25\% and 50\% for the holographic images, with the intention of overcoming the performance limitations posed by the modality difference experienced when shifting from the RGB optical images to the greyscale holographic images, as by expanding bounding boxes, we accounted for possible misalignment between holographic and optical images. Expanding the bounding box areas any further than that would result in microscopic objects other than pollen grains getting detected, which is undesirable.

\begin{figure*}[!htbp]
    \centering
    \begin{subfigure}[b]{0.48\textwidth}
        \centering
        \includegraphics[width=0.79\textwidth]{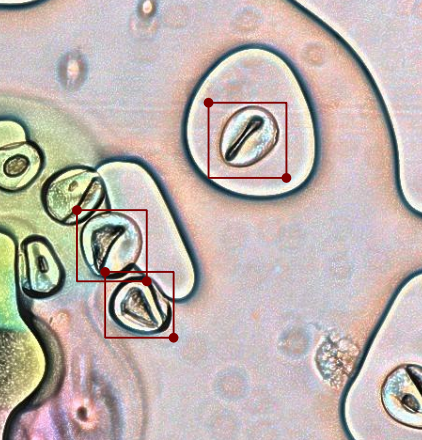}
        \caption{Pollens in an RGB optical image with manual labels and no bounding box expansion.}
        \label{fig:optical-00-expansion}
    \end{subfigure}
    \hfill
    \begin{subfigure}[b]{0.48\textwidth}
        \centering
        \includegraphics[width=0.75\textwidth]{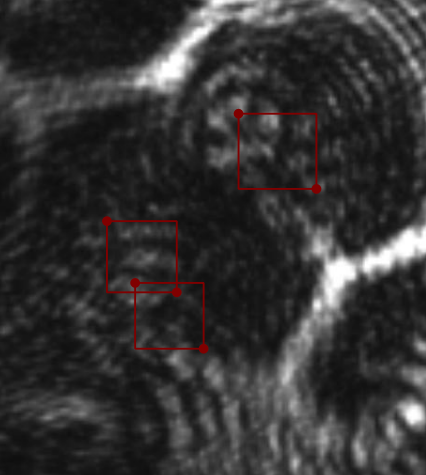}
        \caption{Pollens in an aligned greyscale holographic image with manual labels and no bounding box expansion.}
        \label{fig:holographic-00-expansion}
    \end{subfigure}
    \hfill
    \begin{subfigure}[b]{0.48\textwidth}
        \centering
        \includegraphics[width=0.75\textwidth]{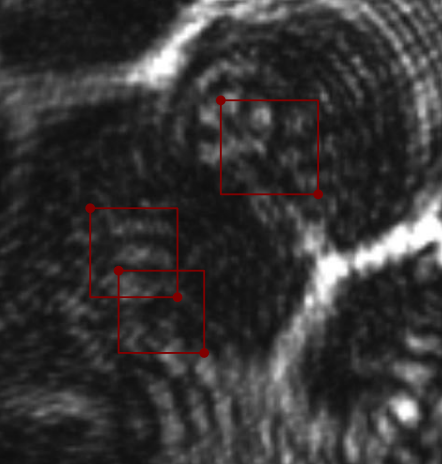}
        \caption{Pollens in an aligned greyscale holographic image with manual labels and 25\% bounding box expansion.}
        \label{fig:holographic-25-expansion}
    \end{subfigure}
    \hfill
    \begin{subfigure}[b]{0.48\textwidth}
        \centering
        \includegraphics[width=0.75\textwidth]{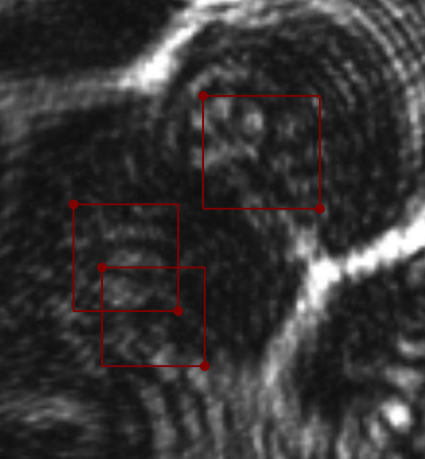}
        \caption{Pollens in an aligned greyscale holographic image with manual labels and 50\% bounding box expansion.}
        \label{fig:holographic-50-expansion}
    \end{subfigure}
    \caption{Comparison of pollens in an optical and an aligned holographic image with manual labels at 94\% zoom.}
    \label{fig:comparison-images}
\end{figure*}

\subsection{Detection Training}

For detecting pollen grains in microscopy images, we utilized the YOLOv8s architecture, pre-trained on the COCO dataset \cite{b19}, chosen for its detection accuracy and inference speed. The high-resolution images were split into smaller 640 by 640 pixels pieces for detection training. We used a one-phase training approach with 200 epochs, uniform learning rate adjustment and an early stopping patience value of 25 epochs. This was because detection tasks benefit from longer training periods with gradual learning rate reduction. We used limited but effective image transformations (see below) to preserve spatial relationships when augmenting images. The class imbalance was handled by the built-in focal loss components and composite loss function of YOLOv8s. The other settings were kept as defaults. We used the mean average precision (mAP) with an intersection-over-union (IOU) threshold of 0.5 as our primary performance metric. 

We implemented the following explicit parameter choices during the object detection model training step to improve performance resilience for novel cases:
\begin{enumerate}
    \item Random Rotation by up to 45 degrees.
    \item Vertical Flipping with a 50\% probability.
    \item Mixup Augmentation, which has a 10\% probability of blending two images.
\end{enumerate}

\subsection{Classification Training}

For classifying pollen grains in microscopy images, we utilized the MobileNetV3L architecture, pre-trained on the ImageNet dataset \cite{b20}, chosen for reasons that are similar to those of the other model. The high-resolution images were split into much smaller pieces with dimensions ranging from 75 pixels to 150 pixels, so the target image size was set to 112 by 112 pixels due to this being exactly half of the default image sizes for the model to maximize compatibility, and all images were adjusted automatically for classification training. We used a two-phase training approach with 30 epochs for transfer learning with a frozen backbone, 30 epochs for fine-tuning with the final 20 layers unfrozen and an early stopping patience value of 10 epochs. This was because dual-phase training provides better effectiveness for fine-grained classification tasks. We used more extensive image transformations (see below) than in case of object detection to boost performance across varying pollen appearances. The class imbalance was handled by an automated class weight computation paired with a weighted loss function. We applied an increased dropout rate of 0.5 to the classifier head of the model for preventing overfitting. We used the overall accuracy of the model as our primary performance metric.

We implemented the following design choices during the object classification model training step to augment the dataset and make the categorical distributions more balanced:
 
\begin{enumerate}
    \item Transformations to augment images.

    \begin{enumerate}
        \item Random Rotation by up to 40 degrees.
        \item Random Horizontal Flipping with a 50\% probability.
        \item Random Affine Transformation, which translates images by up to 20\% in an arbitrary direction.
        \item Random Resized Cropping, which crops and resizes images while maintaining about 80-100\% of the original content.
        \item Intensity/Color Jittering, which randomly adjusts brightness ($\pm$20\%), contrast ($\pm$20\%), saturation ($\pm$20\%), and hue ($\pm$10\%).
    \end{enumerate}
    
    \item Automatic class weight computation: We used an inverse frequency weighting approach by assigning importance weights to each class that are inversely proportional to their frequency in the training dataset. Consequently, minority classes receive substantially higher weights than majority classes.
    
    \item Weighted loss function: We used the cross-entropy loss function with the computed class weights, which increases the penalty for misclassifying samples from minority classes. This makes the model pay greater attention to underrepresented categories during training.
\end{enumerate}

\section{RESULTS}

We report our results separately for the object detection and classification tasks due to the inherent differences in techniques and metrics used for them. 

\begin{figure*}[!htbp]
    \centering
    \begin{subfigure}[b]{0.48\textwidth}
        \centering
        \includegraphics[width=1.0\textwidth]{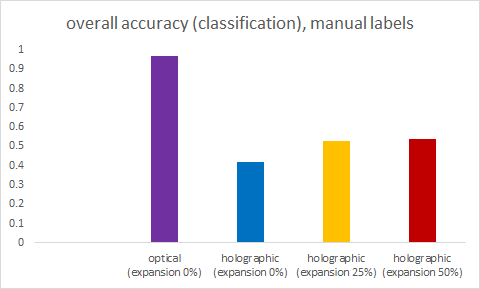}
        \caption{Changes in overall accuracy for MobileNet v3L with manual labels}
        \label{fig:classification-manual}
    \end{subfigure}
    \hfill
    \begin{subfigure}[b]{0.48\textwidth}
        \centering
        \includegraphics[width=1.0\textwidth]{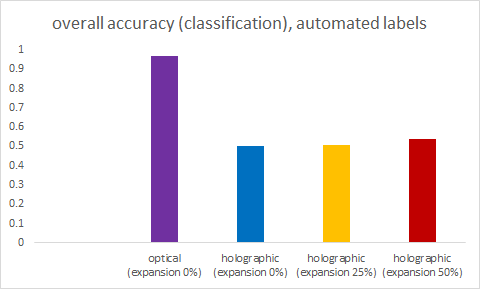}
        \caption{Changes in overall accuracy for MobileNet v3L with automated labels}
        \label{fig:classification-automated}
    \end{subfigure}
    \hfill
    \begin{subfigure}[b]{0.48\textwidth}
        \centering
        \includegraphics[width=1.0\textwidth]{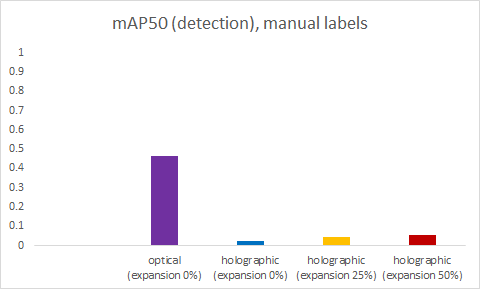}
        \caption{Changes in mAP50 for YOLO v8s with manual labels}
        \label{fig:detection-manual}
    \end{subfigure}
    \hfill
    \begin{subfigure}[b]{0.48\textwidth}
        \centering
        \includegraphics[width=1.0\textwidth]{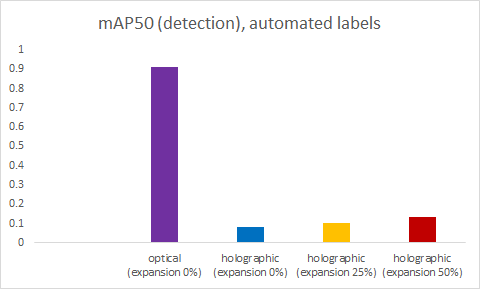}
        \caption{Changes in mAP50 for YOLO v8s with automated labels}
        \label{fig:detection-automated}
    \end{subfigure}
    \caption{Comparison of model performance for bounding box areas of 100\%, 125\% and 150\% with RGB optical and greyscale holographic images}
    \label{fig:comparison-graphs}
\end{figure*}

\subsection{Object Detection}

The mAP50 value for YOLO v8s went from 46.2\% to 91.3\% after shifting from manually made labels (containing 5590 object instances in total) to automatically made labels (containing 77268 object instances in total) with RGB optical images, due to an increase of 13.82 times in the effective dataset, which is a 1.97 times improvement (Fig. 3).

The mAP50 value for YOLO v8s degraded by 18.55 times from 46.2\% to 02.49\% after shifting from RGB optical images to greyscale holographic images with manually made labels for both modalities, which implies a domain-based disparity issue resulting in a performance difference. The decrease in the effective training dataset from 5590 object instances to 2952 object instances could have contributed as well to the performance decline.

Continuing the object detection steps with greyscale holographic images, the use of automatically made labels (containing 71639 object instances in total) instead of manually made labels resulted in the increase of the mAP50 value from 02.49\% to 08.15\% with greyscale holographic images, which is a 3.27 times improvement upon using a 24.26 times larger effective training dataset. The expansion of each bounding box area by 25\% resulted in the increase of the mAP50 value from 02.49\% to 04.32\% for manually made labels, which is a 1.73 times improvement.

The combination of automatically made labels (containing 71639 object instances in total) and area expansion by 25\% saw a 2.36 times increase to 10.2\% from 04.32\% in comparison to using only area expansion by 25\% and also a 1.25 times increase to 10.2\% from 08.15\% in comparison to using only automatically made labels. This implies that combining the two methods causes improvement in results for both the dataset expansion method and the bounding box expansion method.

Expanding the bounding box area by 50\% instead of 25\% was able to increase the mAP50 value from 02.49\% to 05.53\% with manually made labels, which is a 2.22 times improvement. Another increase of 1.63 times to 13.3\% from 08.15\% was observed upon pairing area expansion of 50\% with the automatically made labels in comparison to using only the latter without the former.

\subsection{Object Classification}

The overall accuracy for MobileNetV3L was unchanged at 97\% after shifting from manually made labels (containing 4536 object instances in total) to automatically made labels (containing 68268 object instances in total) with RGB optical images, despite an increase of 15.05 times in the effective dataset, which implies performance plateauing (Fig. 3).

The overall accuracy for MobileNetV3L degraded by 2.30 times from 97\% to 42\% after shifting from RGB optical images to greyscale holographic images with manually made labels for both modalities, which implies a domain-based disparity issue resulting in a performance difference. The decrease in the effective training dataset from 4536 object instances to 2437 object instances could have contributed as well to the performance decline.

Continuing the object classification steps with greyscale holographic images, the use of automatically made labels (containing 63018 object instances in total) instead of manually made labels was able to increase the overall accuracy from 42\% to 50\% with greyscale holographic images, which is a 1.19 times improvement upon using a 25.85 times bigger effective training dataset. The expansion of each bounding box area by 25\% was able to increase the overall accuracy from 42\% to 53\% with manually made labels, which is a 1.26 times improvement.

The combination of automatically made labels (containing 63018 object instances in total) and area expansion by 25\% saw a 1.03 times decrease to 51\% from 53\% in comparison to using only area expansion by 25\% and also a 1.02 times increase to 51\% from 50\% in comparison to using only automatically made labels. This implies that using the automatically made labels together with the expanded bounding box areas causes performance improvement for the former method but performance degradation for the latter method.

Expanding the bounding box area by 50\% instead of 25\% was able to increase the overall accuracy from 42\% to 54\% with manually made labels, which is a 1.28 times improvement. An unchanged value of 54\% was observed upon pairing area expansion of 50\% with the automatically made labels in comparison to using only the former, which implies performance stagnation. However, a 1.05 times improvement from 51\% to 54\% was observed in comparison to using only the latter.

\section{CONCLUSIONS}

Our research confirms that there are both challenges and potential in terms of utilizing deep learning models to improve and automate the analysis of pollen grains across different microscopy imaging modalities. It is apparent that quite a large performance gap exists between the RGB optical and greyscale holographic modalities, which implies that performance transfer is a major matter of concern, at least when it comes to the adoption of machine learning solutions in microscopy imaging. On one hand, when trained for optical microscopy images, the two models performed well, yielding 91.3\% mAP50 for detection tasks and 97\% overall accuracy for classification tasks, thus establishing a strong baseline for automated pollen analysis. However, on the other side, the substantial performance drop to 2.49\% mAP50 and 42\% overall accuracy when applying those same models to holographic images reveals the difficulties faced by the deep learning approach due to resolution differences and the substantial information changes due to domain shift between modalities. Although the image alignment process could have made an impact as well, it was most probably not a significant factor. Overall, the results show that automatic object recognition in holographic images by itself is a much harder task in comparison to optical images. Beyond that, in comparison to the drop in classification performance, the drop in detection performance was especially large.

The methods we applied to address these challenges show promising results. By expanding the dataset through automated labeling, we increased the effective object instances for training by factors ranging from 13.82 times to 25.85 times across different tasks. This approach, combined with bounding box expansion by 25\% and 50\%, yielded successively improving results, with the best being 13.3\% mAP50 for detection and 54\% overall accuracy for classification on holographic images. While these improvements are substantial, they also indicate that further research is necessary, as we have yet to overcome the performance gap of 78\% mAP50 for classification and 43\% overall accuracy for detection between optical and holographic modalities in the context of automated analysis in microscopy imaging. We also need to understand why the drop in detection performance is much greater than the drop in classification performance. 
Future work should focus on the development of training pipelines or model architectures that can better accommodate characteristic features of greyscale holographic images. It is also possible to build upon our approach to improve domain adaptation techniques.

Despite the above limitations, our findings demonstrate that combining lensless digital holographic microscopy with deep learning holds significant potential for expanding the availability of diagnostic facilities in veterinary practice and other fields of medicine. The dramatically lower cost and simplified setup of lensless DHM in comparison to conventional optical microscopy would allow microscopic analysis capabilities to reach underserved regions. They could provide a boost to patient care by cancelling out the limitations of resource-constrained environments. With further maturation of deep learning use in holographic microscopy image processing, we expect the performance gap with traditional microscopy to narrow, making this approach increasingly viable for real-world diagnostic applications.

\section*{Acknowledgment}

This research is funded by the Latvian Council of Science, project VetCyto, project Nr. lzp-2023/1-0220.


\begin{thebibliography}{00}

\bibitem{b1} S. Amann, M. von Witzleben, and S. Breuer, ``3D-printable portable open-source platform for low-cost lens-less holographic cellular imaging,'' Scientific Reports, vol. 9, no. 11260, 2019. 

\bibitem{b2} B. Cugmas, M. Tamosiunas, K. G. Zviedris, M. Bürmen, E. Štruc, L. Goldberga, R. Kadikis, P. Naglič, and M. Ivanovs, ``Automated Classification of Pollens Relevant to Veterinary Medicine,'' 2024 IEEE 14th International Conference Nanomaterials: Applications and Properties (NAP), Riga, Latvia, pp. 103-107, 2024.

\bibitem{b3} M. K. Kim, ``Principles and techniques of digital holographic microscopy,'' SPIE Reviews, vol. 1, p. 018005, 2010.

\bibitem{b4} M. Rodríguez-Damián, E. Cernadas, and A. Formella, ``Pollen classification using brightness-based and shape-based descriptors,'' 2004. 

\bibitem{b5} A. B. Gonçalves, J. S. Souza, G. G. da Silva, M. P. Cereda, A. Pott, M. H. Naka, and H. Pistori, ``Feature Extraction and Machine Learning for the Classification of Brazilian Savannah Pollen Grains,'' PLOS ONE, vol. 11, no. 6, 2016. 

\bibitem{b6} G. Astolfi, A. B. Gonçalves, G. V. Menezes, F. S. B. Borges, A. C. M. N. Astolfi, E. T. Matsubara, M. Alvarez, and H. Pistori, ``POLLEN73S: An image dataset for pollen grains classification,’’ Ecological Informatics, vol. 60, p. 101165, 2020.

\bibitem{b7} R. Gallardo, C. J. García-Orellana, H. M. González-Velasco, A. García-Manso, R. Tormo-Molina, M. Macías-Macías, and E. Abengózar, ``Automated multifocus pollen detection using deep learning,’’ Multimedia Tools and Applications, 2024.

\bibitem{b8} M. A. Rostami, B. Balmaki, L. A. Dyer, J. M. Allen, M. F. Sallam, and F. Frontalini, ``Efficient pollen grain classification using pre-trained Convolutional Neural Networks: a comprehensive study,’’ Journal of Big Data, vol. 10, no. 151, 2023.

\bibitem{b9} Y. Minowa, K. Shigematsu, and H. Takahara, ``A Deep Learning-Based Model for Tree Species Identification Using Pollen Grain Images,’’ Applied Sciences, vol. 12, no. 24, p. 12626, 2022.

\bibitem{b10} J. A. Grant-Jacob, M. Praeger, R. W. Eason, and B. Mills, ``Single-frame 3D lensless microscopic imaging via deep learning,’’ Optics Express, vol. 30, no. 18, pp. 32621-32632, 2022.

\bibitem{b11} V. Sevillano and J. L. Aznarte, ``Improving classification of pollen grain images of the POLEN23E dataset through three different applications of deep learning convolutional neural networks,’’ PLOS ONE, vol. 13, no. 9, p. e0201807, 2018.

\bibitem{b12} B. Cugmas, E. Štruc, M. Tamosiunas, L. Goldberga, I. Bērziņa, R. Kadiķis, M. Ivanovs, S. S. Warshaneyan, and P. Naglič, ``Comparison of two fixation methods in automated pollen classification on whole slide images,’’ 2024. 

\bibitem{b13} L. MacNeil, S. Missan, J. Luo, T. Trappenberg, and J. LaRoche, ``Plankton classification with high-throughput submersible holographic microscopy and transfer learning,’’ BMC Ecology and Evolution, vol. 21, no. 123, 2021. 

\bibitem{b14} J. E. Brito Carcaño, S. Cuenat, B. Ahmad, P. Sandoz, R. Couturier, G. Laurent, and M. Jacquot, ``Digital holographic microscopy applied to 3D computer microvision by using deep neural networks,’’ EPJ Web of Conferences, vol. 287, p. 13011, 2023.

\bibitem{b15} Y. Jo, S. Park, J. Jung, J. Yoon, H. Joo, M. Kim, S. Kang, M. Choi, S. Lee, and Y. Park, ``Holographic deep learning for rapid optical screening of anthrax spores,’’ Science Advances, vol. 3, 2017.

\bibitem{b16} A. Murovec, B. Cugmas, E. Štruc, M. Tamošiūnas, M. Bürmen, and P. Naglič, ``Investigation of Lensless On-Chip Microscopy for High-Fidelity Hologram Reconstruction and Visualization of Pollen Samples,’’ in 2024 IEEE 14th International Conference ``Nanomaterials: Applications and Properties’’, Riga, Latvia, 2024, abstract no. 01bp-17, p. 24.

\bibitem{b17} Ultralytics, ``YOLOv8: A state-of-the-art real-time object detection model,'' 2023. [Online]. Available: \url{https://github.com/ultralytics/ultralytics}

\bibitem{b18} A. Howard, M. Sandler, G. Chu, L.-C. Chen, B. Chen, M. Tan, W. Wang, Y. Zhu, R. Pang, V. Vasudevan, Q. V. Le, and H. Adam, ``Searching for MobileNetV3,'' in *Proc. IEEE/CVF Int. Conf. Comput. Vis. (ICCV)*, Seoul, Korea (South), Oct. 2019, pp. 1314-1324.

\bibitem{b19} T.-Y. Lin, M. Maire, S. Belongie, J. Hays, P. Perona, D. Ramanan, P. Dollár, and C. L. Zitnick, ``Microsoft COCO: Common Objects in Context,'' in \textit{Proc. European Conference on Computer Vision (ECCV)}, 2014, pp. 740–755.

\bibitem{b20} J. Deng, W. Dong, R. Socher, L.-J. Li, K. Li, and L. Fei-Fei, ``ImageNet: A Large-Scale Hierarchical Image Database,'' in \textit{Proc. IEEE Conference on Computer Vision and Pattern Recognition (CVPR)}, 2009, pp. 248–255.

\end{thebibliography}
\end{document}